\pdfoutput=1
\documentclass[nohyperref]{article}

\usepackage{microtype}
\usepackage{graphicx}
\usepackage{subfigure}
\usepackage{booktabs} 

\usepackage{hyperref}

\usepackage[accepted]{icml2022}

\usepackage{amsmath}
\usepackage{amssymb}
\usepackage{mathtools}
\usepackage{amsthm}

\usepackage[capitalize,noabbrev]{cleveref}

\theoremstyle{plain}

\theoremstyle{definition}

\theoremstyle{remark}

\usepackage[textsize=tiny]{todonotes}

\icmltitlerunning{Self-supervision reduces reliance on visual shortcut features}

\begin{document}

\twocolumn[
\icmltitle{Self-Supervision on Images and Text Reduces \\ 
Reliance on Visual Shortcut Features}




\begin{icmlauthorlist}
\icmlauthor{Anil Palepu}{hst}
\icmlauthor{Andrew L Beam}{dbmi}
\end{icmlauthorlist}

\icmlaffiliation{hst}{Department of Health Sciences and Technology, Harvard-MIT, Cambridge, MA, USA}
\icmlaffiliation{dbmi}{Department of Biomedical Informatics, Harvard Medical School, Boston, MA, USA; CAUSALab, Harvard T.H. Chan School of Public Health, Boston, MA, USA; Department of Epidemiology, Harvard T.H. Chan School of Public Health, Boston, MA, USA}

\icmlcorrespondingauthor{Anil Palepu}{apalepu@mit.edu}

\icmlkeywords{Machine Learning, Self-supervised learning, CLIP, Chest x-rays, Chest radiographs, Medical Imaging, Shortcut Learning, Synthetic Shortcuts, Integrated Gradients, Zero-shot classification, ICML}

\vskip 0.3in
]



\printAffiliationsAndNotice{}  

\begin{abstract}
Deep learning models trained in a fully supervised manner have been shown to rely on so-called ``shortcut" features. Shortcut features are inputs that are associated with the outcome of interest in the training data, but are either no longer associated or not present in testing or deployment settings. Here we provide experiments that show recent self-supervised models trained on images and text provide more robust image representations and reduce the model's reliance on visual shortcut features on a realistic medical imaging example. Additionally, we find that these self-supervised models ``forget" shortcut features more quickly than fully supervised ones when fine-tuned on labeled data. Though not a complete solution, our experiments provide compelling evidence that self-supervised models trained on images and text provide some resilience to visual shortcut features.

\end{abstract}

\section{Introduction}

A \emph{shortcut feature} is a recently introduced term to describe the phenomenon of a machine learning model relying on \emph{unstable}, \emph{spurious}, or otherwise unreliable model inputs \citet{bellamy2022structural}. Shortcut features have a potentially strong association with the label in the training data, but this association is broken or potentially reversed in testing or deployment environments. Moreover, shortcut features are often non-causal features that humans would not identify as being important to the prediction task. A provocative example was provided in the context of medical imaging by \citet{geirhos2020shortcut}. In this example, a deep learning model trained to detect pneumonia in chest X-rays (CXRs) relied on \emph{watermarks} indicating the hospital where the patient was seen instead of lung pathophysiology as a radiologist would. This reliance is a result of the watermarks containing \emph{statistically} relevant information since each hospital treats unique patient populations with different baseline risks for pneumonia (\emph{e.g.}, acute care vs. ambulatory settings). However, this association is broken if deployed to hospitals with new watermarks or hospitals that do not use watermarks at all. Shortcut features thus pose a serious challenge to the safe deployment of these models in high-stakes medical settings.

In this work, we examine the robustness of deep learning models to watermark shortcuts on CXRs. We assess the sensitivity of traditional, fully supervised models commonly used in the medical literature to that of recently introduced self-supervised models trained jointly on images and text. Specifically, we consider text-image alignment models such as CLIP \cite{radford2021learning, zhang2020contrastive}, since we hypothesized that this alignment might result in the model ignoring visual features that lacked a corresponding text anchor. 

\section{Data}

\subsection{Datasets}
We used the MIMIC-CXR-JPG \cite{johnson2019mimic} and CheXpert \cite{irvin2019chexpert} datasets, which consist of 227,835 and 224,316 chest radiographs respectively, each with clinical labels for Atelectasis, Cardiomegaly, Consolidation, Edema, and Pleural Effusion. The MIMIC-CXR-JPG dataset, which we utilized for model training, also consists of free text radiology reports corresponding to each CXR. We randomly selected 1\% of the CheXpert training set to be used for model fine-tuning. The set of 234 CheXpert CXRs that were labeled by the consensus of three radiologists was chosen as our test set.  \cite{irvin2019chexpert}

\subsection{Data Preprocessing}
At train time, a series of transformations were applied to the images: random cropping, random horizontal flipping, random affine transformations, color jitter, and gaussian blur. Using these transformations, two data augmentations were produced for each input image. At validation and test time, these transformations were skipped, and the images were instead simply center-cropped to 224 by 224 pixels.

When available, the ``Findings" and ``Impressions" sections from the MIMIC-CXR-JPG radiology reports were extracted and concatenated to produce shortened clinical text. When these sections were not available, we attempted to extract the ``Comparisons" section instead, and in the rare case when that was also missing, we simply extracted the original radiology report and clipped it to 100 words.

\subsection{Injection of Synthetic Shortcuts Using Watermarks}
The original CXRs were corrupted by synthetic shortcuts by adding label-associated watermarks. A unique symbol was assigned to each of the five labels, and at train time, each image was selected to be corrupted by shortcut features with a probability of 0.9. If selected to be corrupted, the image was given ``correct" shortcuts with a probability of 0.9, meaning all positive label symbols were watermarked onto the image. Otherwise (with a probability of 0.1), the corrupted image was given ``incorrect" shortcuts and all negative label symbols were watermarked onto the image. 

Two test sets were generated from the base CheXpert test set for each label. The \emph{Shortcut} test sets had the positive label-specific symbol watermarked onto all label-\emph{positive} images, while the \emph{Adversarial} test sets had the positive label-specific symbol watermarked onto all label-\emph{negative} images. As a result, if a model is overly reliant on the watermarked shortcut features, it should have very high accuracy on the \emph{Shortcut} test sets and very low accuracy on the \emph{Adversarial} test sets. Furthermore, such a model could also perform poorly on the original, uncorrupted CXRs.

\section{Model Details}
\subsection{Contrastive Language-Image Pretraining}
We adopted the CLIP architecture \cite{radford2021learning} to jointly train a vision and text encoder on image-text pairs from MIMIC-CXR-JPG. For our vision encoder, we used a ResNet-50 \cite{ren2016deep} backbone that was pretrained on ImageNet \cite{deng2009imagenet}, and for our text encoder, we used BiomedVLP-CXR-BERT-specialized \cite{boecking2022making}, a text transformer pretrained on radiology reports. A dense layer of size 128 was added as the output to each encoder. All but this final  layer of the text encoder were frozen during training, while the vision model was fully unfrozen.

To train the model, the image and text samples were processed by the encoders to produce their respective embeddings. After computing the cosine similarities between each possible image and text embedding combination, the contrastive loss was computed as the cross entropy between these similarities and the idealized perfect matching of positive image-text pairs. Because we produced two image augmentations from each image, we summed the contrastive loss between each augmentation and the text, as well as between the augmentations themselves \cite{li2021supervision}. We trained for 50 epochs with a learning rate of 0.0001 and batch size of 32, saving the model version that minimized validation loss.

Using the above methodology, we trained one version of the CLIP model on original MIMIC\_CXR images and another on the synthetic data containing watermark-based shortcuts.

\subsection{CNN Training and Fine-tuning}
For our CNN models, we used the same ResNet-50 architecture as the self-supervised model, except we added an output layer with five classification heads corresponding to each of the five clinical labels. To train these models, we computed and summed the binary cross-entropy loss for each label with the same hyperparameters as the self-supervised model. We again trained one version of this CNN model on the original MIMIC\_CXR images and one version on the shortcut-corrupted images.

Afterwards, we fine-tuned each of the models using a randomly selected one percent of CheXpert's training set. This fine-tuning was done on uncorrupted CheXpert radiographs for both CNN models and both CLIP models. For the CNNs, this only required training the models further on this new dataset. For the CLIP models, we detached the vision encoder, added a classification layer with 5 heads (thus constructing an identical architecture to the CNN models), and then trained the models. Using this small subset of CheXpert data, these four models were trained for an additional 100 epochs with a 10-fold reduced learning rate.

\section{Model Evaluation}
\subsection{Zero-shot classification}
The vision encoders of the CLIP models were not directly capable of classification, so to produce predictions from these models, we leveraged the text encoder. For each label, we selected a set of 50 MIMIC\_CXR reports from samples that were exclusively positive for that label, processed these reports through the text encoder, and averaged the resulting embeddings to create a label-specific embedding. Our zero-shot predictions were determined by computing the cosine similarity of the input image embedding to these label-specific embeddings. 

Because cosine similarity is in the range -1 to 1, these predictions could not be interpreted as probabilities. Nevertheless, they could still be utilized to evaluate model discrimination for each label by computing areas under the curve (AUC). As seen in Table 1 of the appendix, these zero-shot CLIP classifiers did not achieve the efficacy of the supervised CNN models. However, given the lack of task-specific training, they demonstrated impressive predictive power and were more robust to shortcut features present in the shortcut and adversarial test sets than the fully supervised CNNs.

\begin{figure}[ht]
\vskip 0.2in
\begin{center}
\centerline{\includegraphics[width=\columnwidth]{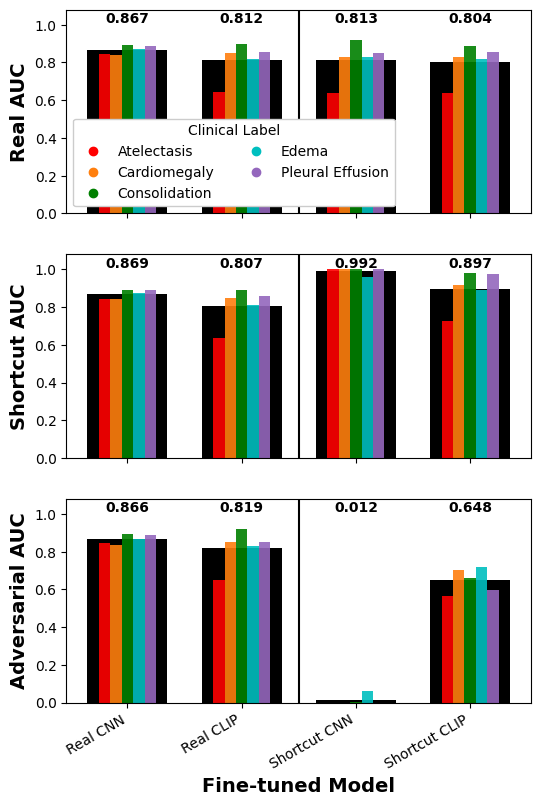}}
\caption{AUC results for the four fine-tuned models on our test sets. These are directly comparable, as the fine-tuned models have identical architectures. The top panel is on the original CXRs, the middle panel is with shortcuts added to the positive label samples, and the bottom panel is with shortcuts added to the negative label samples. These AUCs were computed for each of the five clinical labels, which were averaged to compute the black bars.}
\label{AUCresults}
\end{center}
\vskip -0.2in
\end{figure}

\subsection{Classification Performance}
After fine-tuning, we had four identical CNN architectures that differed only in the initial training strategy (CNN vs CLIP) and the presence of shortcuts in the training data (Real vs Shortcut). We evaluated each of these models by computing the AUC for the five clinical label predictions on the original, shortcut, and adversarial CheXpert test sets. These results can be seen in Figure 1. 

As expected, the models that were trained on real data were largely unaffected by the presence of watermarks in the testing data. In terms of AUC, the fine-tuned CLIP models performed comparably to or slightly better than the fine-tuned CNN models.

The models that were trained on shortcut data performed slightly worse on real data than their uncorrupted counterparts. On the label-associated shortcut test data, the CLIP and CNN models both saw a drastic rise in performance, while on adversarial test data, they both saw a drastic drop in performance. This relationship was especially striking in the CNN model, which achieved near-perfect discrimination (AUC=0.992) on associated shortcuts, and completely failed (AUC=0.012) on adversarial shortcuts. While the CLIP model did learn to use these shortcuts, our results suggest it was significantly less reliant on them and was often able to make predictions in conflict with the shortcuts present.

\begin{figure*}[ht]
\vskip 0.2in
\begin{center}
\centerline{\includegraphics[width=\textwidth]{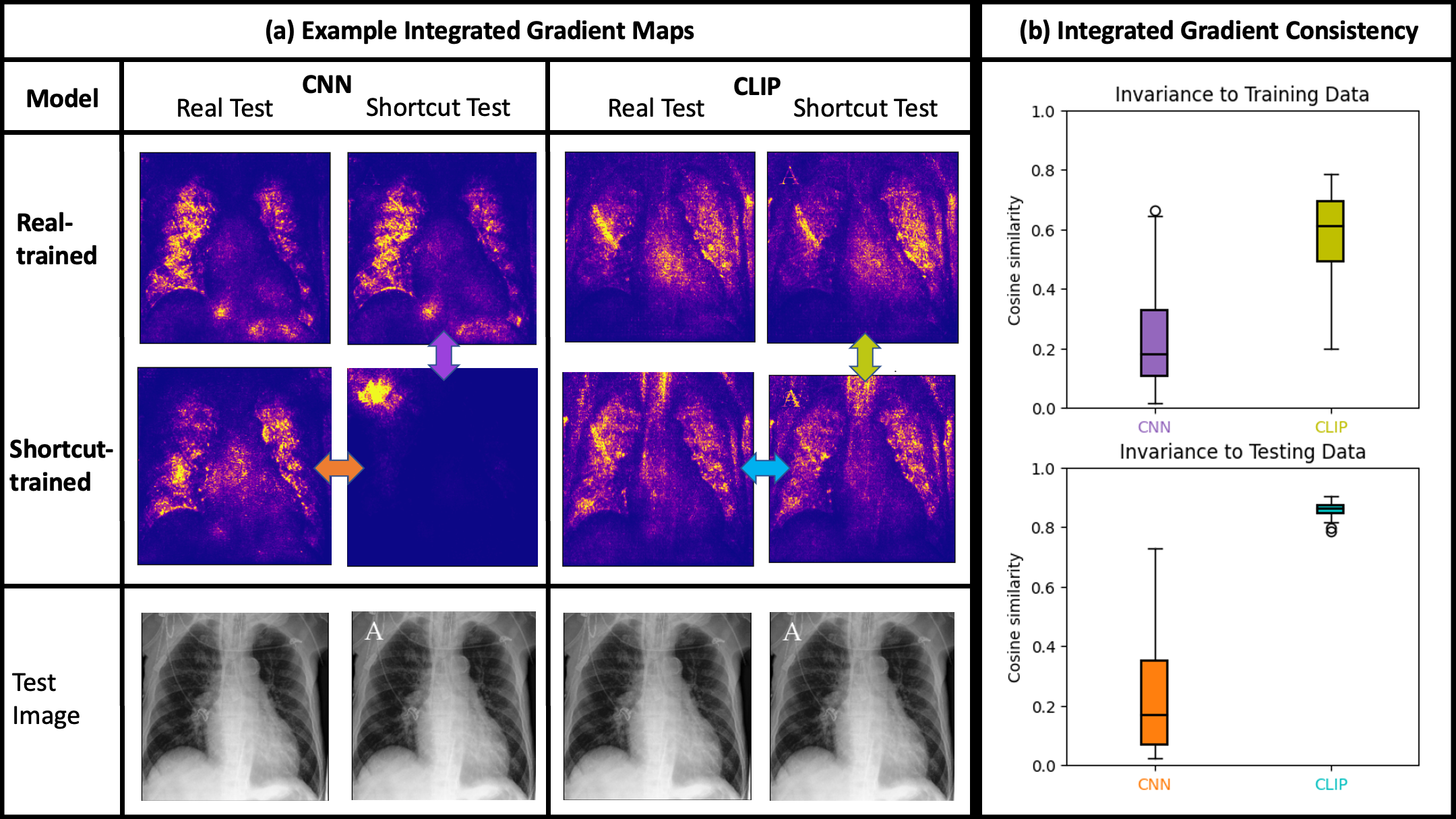}}
\caption{(a) On the left is an example of the integrated gradient maps for Atelectasis prediction using the CNN models prior to fine-tuning. On the right are the corresponding maps using the CLIP models. The colored arrows correspond to the comparisons made in Figure 2(b). (b) After computing the maps in (a) for all CheXpert test images, we computed the pixel-level cosine similarities between maps from the real-trained and shortcut-trained models (top plot) as well as between real and shortcut test images (bottom plot). A higher cosine similarity indicates more consistency between the integrated gradient maps. These plots demonstrate that the CLIP models are more consistent regardless of the shortcuts present at train or test time.}
\label{intgradient}
\end{center}
\vskip -0.2in
\end{figure*}

\subsection{Integrated Gradient Maps}
Integrated gradients are a model attribution method that we leveraged to visualize the features being used by our models \cite{sundararajan2017axiomatic}. The integrated gradients were computed by generating a linear interpolation between a black baseline and the input and accumulating the gradients of the output with respect to these interpolated inputs. The integrated gradient maps on all test images were computed for the models prior to fine-tuning. To do so, we used Google's PAIR saliency package to implement \emph{Integrated Gradient + SmoothGrad}. For each input image, \emph{SmoothGrad} produces several noisy copies of that image and then averages the resulting integrated gradients from each copy \cite{smilkov2017smoothgrad}. 

A key axiom that integrated gradients are designed to fulfill is ``implementation invariance," meaning that functionally equivalent networks should have identical attributions \cite{sundararajan2017axiomatic}. As seen in Figure 2(b), the fact that the CNN models had poor invariance to training and testing data suggests that these models were far from functionally equivalent. For a CNN model trained on shortcuts, the presence of shortcuts drastically altered that model's behavior. In particular, as demonstrated qualitatively by Figure 2(a), the shortcut-trained CNN appeared to exclusively utilize the watermarked shortcuts (when available) to make its predictions.

On the other hand, the CLIP model exhibited far more consistency. The shortcut-trained CLIP had highly similar integrated gradients regardless of whether or not a shortcut was present at test time. Additionally, its integrated gradients were quite similar to those produced by its real-trained counterpart, indicating that these models were functionally similar despite one being trained on shortcut-corrupted data. 

\section{Discussion}
\subsection{Conclusions}
Our results suggest that the natural language supervision provided by the CLIP architecture reduced model reliance on shortcut features. Unlike the CLIP model, the CNN trained on label-associated shortcuts was completely impaired by adversarially manipulating these shortcuts, and was unable to unlearn the shortcuts even with further training on uncorrupted data. The consistency of the integrated gradient maps further corroborate the notion that pretraining with the CLIP architecture allowed the vision encoder to remain relatively robust to shortcuts.

We can explain these results by considering the source of shortcut learning. Shortcuts are learned due to associations that happen to be present in the training data. By forcing the image features to align with a more nuanced text embedding instead of simple binary labels, we can diminish these spurious associations that give rise to shortcut learning and learn more robust visual features. 

\subsection{Limitations}
In this study, we generated shortcut datasets by watermarking label-associated symbols onto the images. This is fairly realistic for CXRs; prior studies have shown that watermarked laterality markers \cite{degrave2021ai} and hospital tokens \cite{geirhos2020shortcut} posed risks for shortcut learning in CXRs. However, the associations we constructed were likely stronger than would be seen in a typical dataset. Nevertheless, using these strongly-correlated shortcuts enabled us to better visualize and manipulate shortcut usage, and we believe our findings will generalize to scenarios with less exaggerated associations.

Even after fine-tuning, the shortcut-trained CLIP model is still somewhat ineffective on adversarial examples, failing almost half of the time. Additionally, while the integrated gradient results suggest that the CLIP model is more consistent in the presence of shortcuts, more investigation is needed to evaluate if these learned features are clinically relevant. 

\section*{Software and Data}

The datasets used are readily available: \\ MIMIC-CXR-JPG: \url{https://physionet.org/content/mimic-cxr-jpg/2.0.0/}, \\ CheXpert: \url{https://stanfordmlgroup.github.io/competitions/chexpert/}. Generation of the synthetic data is done at train/test time and these methods are fully described in the Data section. 

Our code is available here: \url{https://github.com/apalepu13/CNN_CLIP_Shortcut_Feature_Reliance}

\section*{Acknowledgements}

Research reported in this workshop was supported by the National Institute of Biomedical Imaging and Bioengineering (NIBIB), of the National Institutes of Health. The content is solely the responsibility of the authors and does not represent the official views of the National Institutes of Health. ALB received funding from the National Institutes of Health (K01 HL141771).

\bibliography{my_paper}
\bibliographystyle{icml2022}
\newpage
\appendix
\onecolumn
\section{Appendix}
Below are the AUCs for the eight models: (CNN/CLIP) trained on (Real/Shortcut) data (prior to/after) fine-tuning. The CLIP models prior to fine-tuning are performing zero-shot classification as described in the paper, while the rest are directly predicting the five clinical labels. Results are reported on the Real, Shortcut, and Adversarial test sets.

\begin{tabular}{lcccr}
\toprule
Real Test AUCs & Average & Atelectasis/Cardiomegaly/Consolidation/Edema/Pleural Effusion \\
\midrule
Real CNN & 0.866 & 0.807 / 0.840 / 0.895 / 0.902 / 0.886  \\
Zero-shot Real CLIP & 0.749 & 0.556 / 0.841 / 0.776 / 0.853 / 0.720\\
Fine-tuned Real CNN & 0.867 & 0.844 / 0.841 / 0.892 / 0.872 / 0.888\\
Fine-tuned Real CLIP & 0.812 & 0.642 / 0.850 / 0.899 / 0.819 / 0.853     \\
Shortcut CNN  & 0.872 & 0.867 / 0.838 / 0.922 / 0.864 / 0.868 \\
Zero-shot Shortcut CLIP & 0.686 & 0.565 / 0.766 / 0.567 / 0.884 / 0.649\\
Fine-tuned Shortcut CNN & 0.814 & 0.638 / 0.830 / 0.917 / 0.830 / 0.851   \\
Fine-tuned Shortcut CLIP & 0.804 & 0.637 / 0.830 / 0.885 / 0.816 / 0.853\\
\bottomrule
\end{tabular}

\begin{tabular}{lcccr}
\toprule
Shortcut Test AUCs & Average & Atelectasis/Cardiomegaly/Consolidation/Edema/Pleural Effusion \\
\midrule
Real CNN & 0.866 & 0.806 / 0.843 / 0.886 / 0.905 / 0.887 \\
Zero-shot Real CLIP & 0.765 & 0.568 / 0.843 / 0.818 / 0.866 / 0.728\\
Fine-tuned Real CNN & 0.869 & 0.845 / 0.842 / 0.893 / 0.876 / 0.889 \\
Fine-tuned Real CLIP & 0.807 & 0.637 / 0.847 / 0.888 / 0.808 / 0.856  \\
Shortcut CNN  & 1.000 & 0.999 / 1.000 / 1.000/ 1.000 / 1.000\\
Zero-shot Shortcut CLIP & 0.946 & 0.893 / 0.956 / 0.979 / 0.998 / 0.903\\
Fine-tuned Shortcut CNN &  0.992 & 1.000 / 1.000 / 1.000 / 0.960/ 1.000  \\
Fine-tuned Shortcut CLIP & 0.897 & 0.726 / 0.919 / 0.978 / 0.891 / 0.974\\
\bottomrule
\end{tabular}

\begin{tabular}{lcccr}
\toprule
Adversarial Test AUCs & Average & Atelectasis/Cardiomegaly/Consolidation/Edema/Pleural Effusion \\
\midrule
Real CNN & 0.865 & 0.807 / 0.830 / 0.902 / 0.897 / 0.885 \\
Zero-shot Real CLIP & 0.737 & 0.547 / 0.837 / 0.752 / 0.837 / 0.713\\
Fine-tuned Real CNN & 0.866 & 0.845 / 0.837 / 0.894 / 0.867 / 0.886 \\
Fine-tuned Real CLIP & 0.819 & 0.652 / 0.850 / 0.918 / 0.828 / 0.851     \\
Shortcut CNN  & 0.001 & 0.002 / 0.000 / 0.001 / 0.000/ 0.000\\
Zero-shot Shortcut CLIP & 0.165 & 0.111 / 0.356 / 0.036 / 0.224 / 0.098 \\
Fine-tuned Shortcut CNN &  0.012 & 0.000 / 0.000 / 0.001 / 0.060 / 0.000  \\
Fine-tuned Shortcut CLIP & 0.648 & 0.567 / 0.701 / 0.659 / 0.719 / 0.595\\
\bottomrule
\end{tabular}

\end{document}